%% file: main.tex
\documentclass[conference]{IEEEtran}
\pdfoutput=1
\usepackage{cite}
\usepackage{amsmath,amssymb,amsfonts}
\usepackage{algorithmic}
\usepackage{graphicx}
\usepackage{textcomp}
\usepackage{xcolor}

\usepackage{forest}
\usepackage{nccmath}
\usepackage{subfigure}
\usepackage{amsmath}
\usepackage{boldline}
\usepackage{float}

\def\BibTeX{{\rm B\kern-.05em{\sc i\kern-.025em b}\kern-.08em
    T\kern-.1667em\lower.7ex\hbox{E}\kern-.125emX}}

\begin{document}

\title{Adaptation of MobileNetV2 for Face Detection on Ultra-Low Power Platform}

\author{
\IEEEauthorblockN{Simon Narduzzi}
\IEEEauthorblockA{\textit{Edge AI and Vision Group} \\
\textit{CSEM}\\
Neuchâtel, Switzerland \\
simon.narduzzi@csem.ch}
\and
\IEEEauthorblockN{Engin Türetken}
\IEEEauthorblockA{\textit{Edge AI and Vision Group} \\
\textit{CSEM}\\
Neuchâtel, Switzerland \\
engin.tueretken@csem.ch}

\and
\IEEEauthorblockN{Jean-Philippe Thiran}
\IEEEauthorblockA{\textit{Signal Processing Laboratory 5} \\
\textit{EPFL}\\
Lausanne, Switzerland \\
jean-philippe.thiran@epfl.ch}

\and
\IEEEauthorblockN{L. Andrea Dunbar}
\IEEEauthorblockA{\textit{Edge AI and Vision Group} \\
\textit{CSEM}\\
Neuchâtel, Switzerland \\
andrea.dunbar@csem.ch}

}

%\name{Simon Narduzzi$^{\star}$ \qquad Engin Türetken$^{\star}$ \qquad Jean-Philippe Thiran$^{\dagger}$ \qquad L. Andrea Dunbar$^{\star}$}
%\address{\small{$^{\star}$Centre suisse d'électronique et de microtechnique (CSEM), Switzerland} \qquad 
%\small{$^{\dagger}$Ecole polytechnique fédérale de Lausanne (EPFL), Switzerland}}

\maketitle

\begin{abstract}
Designing Deep Neural Networks (DNNs) running on edge hardware remains a challenge. Standard designs have been adopted by the community to facilitate the deployment of Neural Network models. However, not much emphasis is put on adapting the network topology to fit hardware constraints. In this paper, we adapt one of the most widely used architectures for mobile hardware platforms, MobileNetV2, and study the impact of changing its topology and applying post-training quantization. We discuss the impact of the adaptations and the deployment of the model on an embedded hardware platform for face detection.

\end{abstract}


\input{tex/introduction}
\input{tex/related_work_new}
\input{tex/datasets}
\input{tex/methods}
\input{tex/results}
\input{tex/k210}
\input{tex/conclusion}

%\section*{Acknowledgment}

%\section*{References}

\input{biblio}

%\begin{thebibliography}{00}

%\end{thebibliography}
\vspace{12pt}
\color{red}

\end{document}

%% file: tex/introduction.tex
\section{Introduction}
Facial analytics and recognition is a rapidly growing market, with applications ranging from surveillance and access control systems in smart buildings to retail stores that collect viewership and demographics. A major barrier to rapid adaptation of the technology is the computationally demanding nature of the algorithms. In recent years, the progress in artificial intelligence, and more specifically in deep learning, has been providing impressive results in a large variety of tasks, from object detection and face detection to natural language processing and music generation. The downside of these rapid improvements is that a large part of the deep learning algorithms is replaced by a more accurate or a faster implementation within a few months of deployment in order to keep up with state-of-the-art.

Research and industry have been putting a lot of effort into developing systems that can embed these complex algorithms in specialized chips, with hardware designed for specific fast inference. The large variety of neural network architectures forces manufacturers to create generic hardware that can support a wide range of operations. Intel Movidius and NVidia Jetson X1 are some of the most popular specialized platforms used for the inference of deep neural networks. While these platforms maintain a good accuracy of the deployed algorithms, they still require a lot of energy and contain large hardware components, making them unsuitable for very low-power machine learning running on battery. The Kendryte K210 is another RISC-V dual-core processor, running on 64bits architecture and consuming around 1W in typical application scenarios. It is one of the promising solutions in applications with low-power, small-size, and low-cost requirements.

The experiments presented in this paper aim at developing efficient deep learning algorithms for face detection by transforming and adapting common architectures to embed them into low power and resource constraint mobile platforms, and provide solutions to embed models in devices similar to the K210 processor. The main challenges addressed are finding an efficient architecture for the facial detection problem and optimizing it to maintain reliability in detection accuracy, while minimizing the computational resources required to evaluate the AI model.

The structure of the paper is the following: the first section reviews the state-of-the-art methods for efficient deep learning, object detection tasks, and face detection. Section \ref{sec:datasets} contains the details about the dataset used for the training and evaluation of the deep learning model. Section \ref{sec:methods} explains the algorithms used in the experiments and introduces the notion of heatmaps and clusters of bounding boxes. Lastly, the results of different face detection algorithms and their efficient implementation on a Kendryte K210 are presented in Section~\ref{sec:results}, and Section~\ref{sec:conclusion} concludes the paper.

%% file: tex/related_work_new.tex
\section{Related Work}
\label{sec:related_work}

\subsection{Efficient architectures}
Over the past years, deep learning researchers put a lot of effort into finding efficient architectures to increase the inference throughput of neural networks and reduce the exploitation costs. This section summarizes some of the techniques used to reduce the neural network algorithms' energy by optimizing the number of computations and parameters storage costs. In 2017, a team of researchers introduced the concept of MobileNets~\cite{howard2017mobilenets}, neural networks that make use of \textit{depthwise separable convolutions}, producing light-weight models for embedded applications. The MobileNet framework showed strong results on different use cases, such as Object-Detection, Face Embeddings, and Face Attributes classification tasks. MobileNetV2~\cite{zhu2018mobilenetv2} further improves the state-of-the-art performance of mobile models by introducing \textit{Inverted Residual blocks} and \textit{Linear Bottlenecks}, which reduce the number of multiply-accumulate (MAC) operations of a model quadratically. ShuffleNet~\cite{zhang2018shufflenet} introduces a novel form of group convolutions and depthwise convolutions that shuffle the channels, eventually leading with better accuracy without more parameters. ResNeXt~\cite{xie2017aggregated} makes extensive use of group convolutions, but only in the $3\times3$ convolution layers, which results in pointwise convolutions occupying $93.4\%$ of the MACs.

Automated techniques for architecture search have been developed in order to explore the architecture space more efficiently. NASNet~\cite{zoph2016neural} uses Reinforcement Learning combined with a recurrent neural network to generate convolutional neural networks with skip connections. Other techniques rely on previously validated architecture (ResNet~\cite{he2016deep}, Inception~\cite{szegedy2015going}) and optimize the number of repetition of the same cell types~\cite{cai2018efficient} or to design task-specific architecture, such as Face Recognition~\cite{zhu2019neural}. Recently, platform-aware reinforcement learning was used to develop MobileNetV3~\cite{howard2019searching}.

\subsection{Compression Techniques}
Besides optimizing the architecture by reducing the number of operations computed inside the layers, many other techniques have emerged to minimize the size of neural networks by reducing the storage required by the parameters and allowing the embedding of the architectures in low-resource environments. Quantization is a technique that transforms a word into a smaller bits representation. In neural networks, quantization is an efficient and straightforward way of reducing the size of the model by representing weights and biases in reduced precision words and has many advantages~\cite{krishnamoorthi2018quantizing}, namely, reducing the storage memory, reducing the working memory, eventually leading to faster computations and less power consumption.

Quantization can be applied during or after training. In quantization-aware training, the weights are optimized during the training phase, providing better accuracy than the post-training quantization. Post-training quantization is simpler to apply and allows more flexibility. Once trained, the weights can be represented in any floating-point or binary format at the cost of lower accuracy than quantization-aware ones. Post-training quantization using a uniform symmetric quantizer has been applied to MobileNets\cite{kulkarni2021quantization} performing image classification, allowing the deployment of the models on mobile devices with reduced memory footprint.

Other techniques have been developed to reduce the memory footprint of the parameters used by neural networks. 
Pruning~\cite{lecun1990optimal, karnin1990simple} removes weights that contribute less to the overall performance of the network. HashedNets~\cite{chen2015compressing} cluster the weights and store them using a low-cost hash function to randomly weights into hash buckets,  with all connections in the same bucket sharing the same values. Huffman coding might also be used to encode the weights efficiently~\cite{han2015deep}. Knowledge distillation~\cite{hinton2015distilling, yim2017gift} is a technique that copies the knowledge of a large network into a smaller one and is another strategy for minimizing the size of a model. Finally, the techniques mentioned above can be combined to obtain very efficient networks~\cite{han2015deep}.

\subsection{State-of-the-art in Face Detection}
\label{sec:dl_vision}
Face detection is an important problem in computer vision. Past methods have been relying on cascade structures and deformable part models (DPM)~\cite{yan2014face}. For a long time, the Viola-Jones algorithm~\cite{viola2001rapid}, based on a cascade of classifiers, was considered state of the art for face detection tasks. The algorithm was surprisingly performant given the handcrafted nature of the features used in the classifiers. With the rise of Deep Learning, Convolutional Neural Networks have also been applied to the face detection problem, many of which rely on a cascade detection pipeline. Most of them use object detection-like frameworks, which in this case is reduced to a single-class prediction problem (face / non-face).

While face bounding boxes have been used as standard labeling methods for the detection of faces, Faceness~\cite{yang2017faceness} uses facial attribute annotations to detect faces. Other architectures perform face detection jointly with other tasks, such as landmark detection in~\cite{zhang2016joint}. The authors of Tiny-Face~\cite{hu2017finding} present a novel approach to face detection. Instead of using multiple classifiers, three networks with shared weights are trained for different scales, such that the resulting single-architecture learns the scale-invariant features of the faces. By running an experiment on human classification, the authors showed that the context is particularly helpful in finding small faces, suggesting that context should be modeled in a scale-variant manner.

%% file: tex/datasets.tex
\section{Datasets}
\label{sec:datasets}

For our experiments, we use two datasets that are presented in this section. The first one is WIDER face, a benchmark dataset for face detection and one of the largest public dataset available today. It contains $32203$ images and $393703$ annotated faces with a high degree of scale, variability, pose and occlusion. The dataset is divided into $40\%$, $10\%$ and $50\%$ testing set. The annotations of the testing set are not made publicly available. The annotations contain top-left and bottom-right coordinates of the bounding boxes.

\begin{figure}[ht]
	\centering
	\subfigure{\includegraphics[height=2.2cm, keepaspectratio]{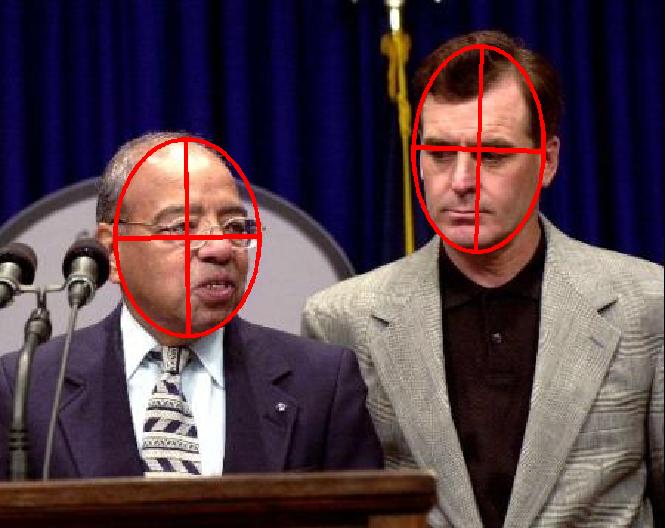}\label{fig:fddb1}}
	\hspace{0.5cm}
	\subfigure{\includegraphics[height=2.2cm, keepaspectratio]{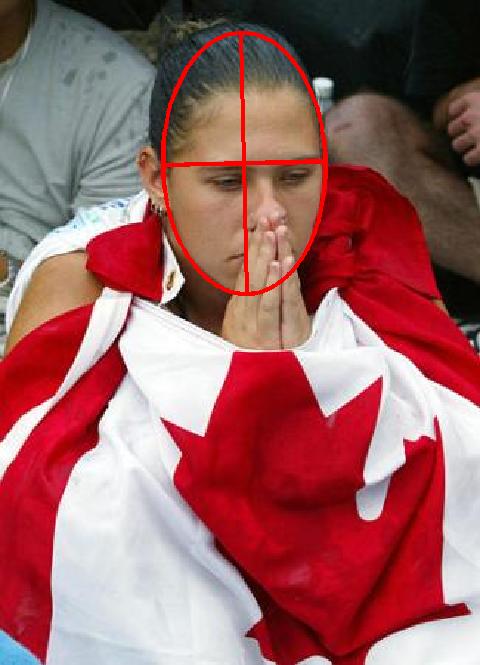}\label{fig:fddb2}}
	\hspace{0.5cm}
	\subfigure{\includegraphics[height=2.2cm, keepaspectratio]{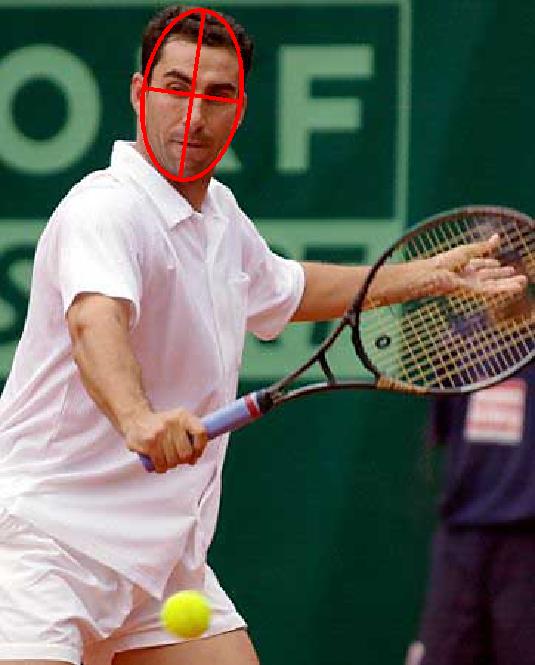}\label{fig:fddb3}}
	\caption{Sample images from FDDB with annotations.}
	\label{fig:fddb_sample}
\end{figure}

 The second dataset is FDDB~\cite{fddbTech}, containing $2845$ images and $5171$ annotated faces in unconstrained conditions. The images of the dataset were originally taken from the Faces in the Wild~\cite{berg2005s} dataset. The annotations contain the parameters of the ellipse covering the face, as shown in Fig.\ref{fig:fddb_sample}. To compare the results with between the datasets, FDDB ellipses are transformed into squared boxes with the following transformation:

\begin{ceqn}
	\begin{align}
	{w}_{box} &= \alpha {w}_{ellipse} &	{h}_{box} &= \alpha {w}_{ellipse} \\
	{c_x}_{{box}} &= {c_x}_{ellipse} &
	{c_y}_{{box}} &= {c_y}_{ellipse} + \beta {w}_{ellipse}
    \end{align}
\end{ceqn}

where $w$,$h$ and $c$ are the width, height and center coordinates respectively and the coefficients $\alpha=1.3$ and $\beta=0.26$ are empirically chosen.

%\subsection{Verification dataset}
%CSEM has created the verification dataset (VD) used to test the algorithms. This dataset is composed of faces extracted from CelebA~\cite{liu2015faceattributes} and transformed to greyscale. The background is cropped around the squared bounding box containing the face with ratio [$1.3$, $3.0$]. Then, the resulting cropped face is randomly resized, rotated and placed on a food background image of size $640\times640$ pixels. Fig. \ref{fig:vd_sample} shows the resulting images.

%\begin{figure}
%	\centering
%	\subfigure{\includegraphics[height=2.0cm, keepaspectratio]{images/sample_ver1}\label{fig:ver1}}
%	\subfigure{\includegraphics[height=2.0cm, keepaspectratio]{images/sample_ver2}\label{fig:ver2}}
%	\subfigure{\includegraphics[height=2.0cm, keepaspectratio]{images/sample_ver3}\label{fig:ver3}}
 %   \subfigure{\includegraphics[height=2.0cm, keepaspectratio]{images/sample_ver4}\label{fig:ver4}}
%	\caption{Sample images from VD with annotations.}
%	\label{fig:vd_sample}
%\end{figure}

%% file: tex/methods.tex
\section{Methods}
\label{sec:methods}
\subsection{Bounding boxes regression}

When a neural network does not use fully-connected layers, it makes it suitable to be used on images of any dimensions. To detect objects in natural images, a typical representation of the neural network's output is based on a 2D-convolution output map containing $n+4$ channels, where $n$ is the number of classes. The most basic output for a single-class detection is a $5$-channels output map. The first channel predicts the class (face) probability, while the remaining channels predict the bounding box's width, height, and displacement of the bounding box's center from the center pixel in the $x$ and $y$ dimensions. This kind of output does not exploit anchor boxes and can be seen as a single-class output map of YOLO \cite{YOLO9000}. The first channel outputs a probability between 0 and 1 of the cell to contain a face (using a sigmoid activation), while the $4$ other channels can take values between $-\infty$ and $+\infty$ (linear activation). A bounding box is represented by a $4$-dimensional vector $t$, which values are computed as follows:

\begin{equation}
\resizebox{0.90\linewidth}{!}{
    $t_x = \frac{x - x_a}{w_a}$\hspace{0.15cm}
	$t_y = \frac{y - y_a}{h_a}$\hspace{0.15cm}
    $t_w = \log\left(\frac{w}{w_a}\right)$\hspace{0.15cm}
	$t_h = \log\left(\frac{h}{h_a}\right)$ 
	}
\end{equation}

where $x$, $y$, $w$ and $h$ denote the coordinates of the box center coordinates, width and height predicted by the network. Variables $x_a$, $y_a$, $w_a$ and $h_a$ are the values associated with the canonical box. We regress these values using a loss function adapted from Faster R-CNN \cite{ren2015faster}:
\begin{equation}
\resizebox{.9\linewidth}{!}{
$L(p,b) = \frac{1}{N_{cls}} \sum_i CE(p_{i}, p^\star_{i}) + \lambda \frac{1}{N_{reg}} \sum_i p^\star_{i}MSE(b_i, b^\star_i)$
}
\end{equation}

where $p_i$ and $b_i$ are the probability and vector representing the bounding box associated with pixel $i$, and $CE$ and $MSE$ are the cross-entropy loss and mean squared error functions, respectively. In Faster R-CNN, $N_{cls}$ represents the number of positive pixels associated with the class, and $N_{reg}$ is the number of pixels of the regression vector where the loss is applied. The purpose of these variables is to better deal with class imbalance. For the sake of simplicity, we fix $N_{cls}=256$ and $N_{reg}=4$.

\subsection{Transfer Learning on MobileNetV2}
We select Tiny-Face as a starting point for developing efficient algorithms running on low-power mobile platforms. The authors computed a set of canonical bounding boxes derived by clustering them into $25$ different sizes. They jointly trained models of $3$ resolutions ($0.5\times$, $1.0\times$, $2\times$, interpolation) to predict the corresponding bounding boxes. During inference, $3$ different scales of the input image are fed to the network that predicts a template response at every resolution. The final detection results are obtained after applying non-maximum suppression (NMS) at the original resolution. For simplicity, we use the same canonical box as provided by the authors.

Inspired by Tiny-Face, we adapted a MobileNetV2 architecture to detect faces. MobileNetV2 is a very light network with separable convolutions. Instead of $3$ networks, we sequentially reuse the same architecture on $3$ different scales by adapting the network to a fully convolutional architecture. Initially built for classification purposes, the network output (and global average pooling followed by a dense layer) is replaced by the $5$-channels 2D convolution. We set the input dimension for training to $500\times500\times3$. To keep the output dimensions as close to the original Tiny-Face network, we early-exit on breakpoints defined at the following locations:
\begin{itemize}
\item Breakpoint \textit{A}: located at layer \#$119$: the output is a tensor of shape $32\times32\times576$
\item Breakpoint \textit{B}: located at layer \#$154$: the last layer, just before the global average pooling. The output shape is $16\times16\times1280$
\end{itemize}

These two breakpoints are used to create three different outputs:
\begin{itemize}
\item \textit{OutA}: Append the output convolution block after breakpoint \textit{A}.
\item \textit{OutB}: Append the output convolution block after breakpoint \textit{B}.
\item \textit{OutC}: Concatenating the breakpoint outputs \textit{A} and \textit{B} (upsampled $2\times$ with bilinear interpolation) into a $32\times32\times1856$ tensor, and then appending the output convolution block.
\end{itemize}

\begin{figure}[h!]
    \centering
    \includegraphics[width=1.0\linewidth]{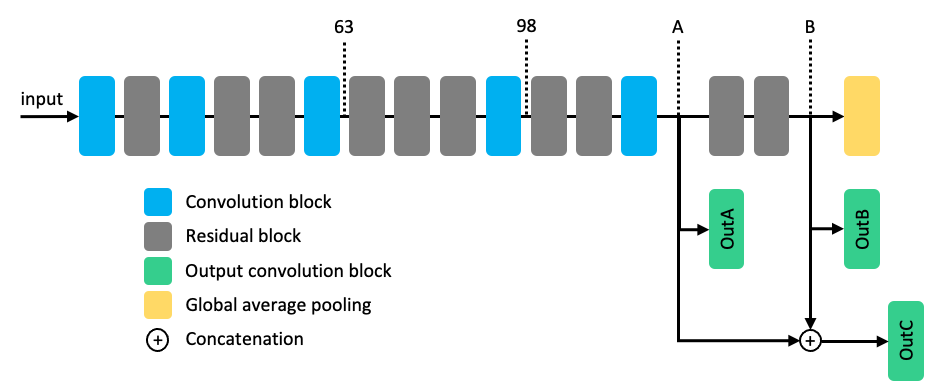}
    \caption{Schematic view of the adapted MobileNetV2 architecture.}
    \label{fig:mbv2}
\end{figure}

The network was initialized with ImageNet weights and fine-tuned on WIDER FACE with two different settings: given a layer index $\ell$, we fix the network weights until that layer. In the experiments, $\ell$ takes the values $63$, $98$, and $133$, corresponding to the input layers of the 4th and 8th residual blocks of the MobileNetV2 architecture. All the models were fine-tuned for $200$ epochs with a learning rate of $1e-4$. Figure \ref{fig:mbv2} shows the experimental setup.

\subsection{Quantization in Q-Format}
While different quantization schemes are available, some models have been quantized down to ternary \cite{hwang2014fixed} and binary networks \cite{rastegari2016xnor}. Here, we explore a specific type of quantization (Q-Format \cite{oberstar2007fixed}). The reason for this choice is influenced by the target platform that will be presented in the following sections. A Q number is composed of an integer part and a fractional part: Q$m.n$ or Q$n$ represent a number where $m$ bits are used for the integer part, and $n$ bits are used for the fractional part. Such number requires $n+m+1$ bits for storage. The choice of $n$ is a tradeoff between the dynamic and the maximal range of the numbers represented. The conversion between Q numbers and floating-point numbers is defined as follows:

\begin{ceqn}
	\begin{align}
	x_{integer} = \lfloor x_{float} \rfloor*2^{n-1}
	\end{align}
\end{ceqn}

\begin{ceqn}
	\begin{align}
	x_{float} = \frac{x_{integer}}{2^{n-1}}
	\end{align}
\end{ceqn}

where $n$ is the number of bits used for the precision.

%% file: tex/results.tex
\section{Results}
\label{sec:results}

\subsection{Selection of output strategy}
\label{sec:strategy}

A first experiment on the MobileNetV2 consists in exploring the output strategy. The breakpoint location and the number of non-trainable layers impact the performance and the size of the model: the later we place the output block, the more parameters the model contains, and the smaller the output tensor is. In general, OutA-type outputs perform better than OutB-type outputs (Table \ref{tab:out_effect}). One explanation might reside in the output dimensions: OutA has larger output maps, and each pixel is responsible for predicting a smaller field in the image and therefore have more discriminative power when two faces are next to each other. 

When the network uses OutB-type blocks (with tensors of dimensions $16\times16\times125$), it performs better if only a few layers are retrained. The drop in accuracy when more layers are trained might be caused by overfitting: the last layers contain $576$ layers originally responsible for classifying objects of ImageNet ($1000$ categories). Using those filters to predict the belonging to $25$ precomputed bounding boxes might be a bit too much. Reducing the number of filters in the last layer could lead to better results. Finally, models with OutC-type output blocks obtain good results but are still less performant than OutA-type outputs that have the same resolution. This supports the hypothesis that the network is overfitting when using the OutB-type outputs but might be regularized by the OutA-type outputs. 

As it can be observed in Table \ref{tab:out_effect}, the models tend to perform poorly when a large number of layers are retrained. Training too many layers can cause catastrophic forgetting \cite{kirkpatrick2017overcoming}, which is the fact of destroying the features learned during the pretraining phase and replacing them with features relative to our dataset, impacting generalization. Therefore, training only the last layers limits the risk of catastrophic forgetting. More experiments are necessary to confirm these hypotheses.

%\begin{figure}[H]
%	\centering
%	\includegraphics[width=6cm]{images/mobilenet_results_fddb}
%	\caption[Results of the MobileNetV2 relative to the output type]{Results of the MobileNetV2 depending on number of retrained layers and output type. OutA-type performs generally better than OutB-types and OutC-types.}
%	\label{fig:results_mobilenet}
%\end{figure}

\begin{table}[t]
    \centering
    \begin{tabular}{lcccc}
		\hlineB{3} \\
		\textbf{Model} & \textbf{\begin{tabular}[c]{@{}c@{}}Output \\ type\end{tabular}} & \textbf{\begin{tabular}[c]{@{}c@{}}Fixed\\ layers\end{tabular}} & \textbf{\begin{tabular}[c]{@{}c@{}}Trainable\\ layers\end{tabular}} & \multicolumn{1}{c}{\textbf{\begin{tabular}[c]{@{}c@{}}AP on \\ FDDB-C\end{tabular}}} \\ \hline
		A63 & OutA   & 63     & 58         & 0.8238    \\ 
		A98 & OutA   & 98     & 23& \textbf{0.8915}\\ 
		B63 & OutB   & 63     & 93         & 0.7466    \\ 
		B98 & OutB   & 98     & 58         & 0.6791    \\ 
		B133& OutB	  & 133    & 23 & 0.8044 \\ 
		C98 & OutC	  & 98	   & 58 & 0.8739 \\ 
		\hlineB{3}
		\\
	\end{tabular}
	\caption[Results of the MobileNetV2 relative to the output type]{Results of the MobileNetV2 depending on the number of retrained layers and output type. OutA-type performs generally better than OutB-types and OutC-types.}
	\label{tab:out_effect}
\end{table}

\subsection{Quantization of MobileNetV2}

\begin{figure}[h]
	\centering
	\subfigure{\includegraphics[width=1.0\linewidth, keepaspectratio]{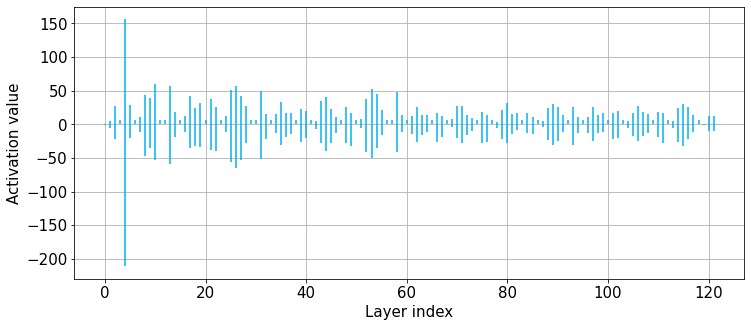}\label{fig:w_min_max}}
	%\hspace{0.5cm}
	%\subfigure{\includegraphics[height=4cm, keepaspectratio]{images/distribution_weights_mobilenet}\label{fig:w_hist}}
	\caption{Activation analysis : Min-max ranges of the activations per layer.}
	\label{fig:w_distribution}
\end{figure}

We then chose the best performing model (A98, $\alpha=1.0$) presented in Section \ref{sec:strategy} and quantized it using Q-format. Because the original design of MobileNetV2 does not contain biases, the first step of the quantization is to plot the distribution of the weights and activation range of each layer. For simplicity, we use a uniform quantization across the layers and hence only look at the min and max value of the total weights and activation distribution. To compute the distribution, $100$ images from the FDDB dataset were chosen randomly and passed to the network to compute the activation distribution. At each layer, the activation map is registered. Figure \ref{fig:w_distribution} shows the activation range of each layer. The maximal activation response occurs at layer $5$, with a range spanning between $-242.14$ and $+155.91$. If we want to encode that maximal range, at least $8$ bits are necessary to encode the integer part. The majority of the activation values are distributed between $-50$ and $+50$. The weights of the network are centered around zero, with a mean value of $0.4583$ and a standard deviation of $2.3385$ across all layers. Most of the weights are distributed between $-8$ and $8$, but layer $5$ spans between $-25$ and $20$. If we want to represent this range, $5$ bits are necessary for the integer part. If we managed to reduce the range of layer $5$ using, for example, regularization or quantization-aware training, we might be able to represent the floating points value with higher precision, as fewer bits will be necessary for integers. Considering the above analysis, the optimal representation of the weights should be 8 bits.
%\begin{figure}[h]
%	\centering
%	\includegraphics[width=7cm]{images/min_max_activations}
%	\caption{Activation analysis : Min-max ranges of the activations per layer.}
%	\label{fig:activation_analysis}
%\end{figure}

Two methods are explored for a simulated quantization: floating-point format on $16$ bits (FP16) and Q-format with different floating-point precision. The reason to convert the weights in Q-format is the following: to save space, some embedded architecture does not include floating-point units, and floating points are therefore simulated using integer computation, such as Q-format. If the architecture contains a floating-point unit, then it can be interesting to compress the weights in a $16$-bits floating-point format to save space and energy consumption. 

The network parameters are converted to a quantized format (Q-format and FP16) and converted back to a $32$-bits floating-point to simulate the network in TensorFlow. Because TensorFlow runs on a $32$-bits system, the intermediate activations might contain values that should have overflowed on a $16$-bits system. Assuming no overflow in activations, Table \ref{tab:quant_effect} shows the effect of post-training quantization on the performance of the A98 network.

%\begin{figure}[h!]
 %   \centering
  %  \includegraphics[width=7cm]{images/quantization_FPR}
   % \caption{TPR/FPR curve for the different quantization of A98}
    %\label{fig:quant_effect}
%\end{figure}

\begin{table}[h]
    \centering
    \begin{tabular}{lcccc}
			\hlineB{3}\\
			\multicolumn{1}{c}{\textbf{Quantization}} & \multicolumn{1}{c}{\textbf{\begin{tabular}[c]{@{}c@{}}AP\end{tabular}}} & \multicolumn{1}{c}{\textbf{Parameters}} & \multicolumn{1}{c}{\textbf{Size}} \\ \hline
			FP32 (base)  & \textbf{0.891}     & 1.37M     & 5.25MB\\
            FP16   & \textbf{0.891}     & 1.37M     & 2.63MB\\
           % Q10    & 0.068  				& 1.37M     & 2.63MB\\ 
            Q9     & 0.886    & 1.37M     & 2.63MB   \\
            Q8     & 0.884 			& 1.37M     & 2.63MB\\ 
            Q7     & 0.814 			& 1.37M     & 2.63MB\\ 
            Q6     & 0.027			  & 1.37M     & 2.63MB\\ 
		     \hlineB{3}
		     \\
	\end{tabular}
    \caption{Performance of MobileNetV2 on FDDB relative to the quantization of A98.}
    \label{tab:quant_effect}
\end{table}

\subsection{Performance tradeoffs}

The width multiplier $\alpha$ of MobileNetV2 is a hyper-parameter that adjusts the number of filters in the depthwise convolutions. The width multiplier is applied to all layers except the very last convolutional layers (output blocks). When $\alpha < 1.0$, the number of filters of each convolution is reduced by a ratio $\alpha$. When $\alpha=1.0$, the original number of filters is used. In TensorFlow, pre-trained networks weights are available for $\alpha=1.0$, $\alpha=0.5$ and $\alpha=0.35$. Table \ref{tab:alpha_mbv2} shows the performance of the network relative to the output strategy and compression ratio $\alpha$. Setting the width multiplier to $\alpha=0.5$ in A98 model leads to a performance drop in AP of $5\%$ but reduces the size of the parameters by $4\times$. The performance is still better than B133, with about $12 \times$ fewer parameters. Further reducing $\alpha$ considerably impacts the performance.

%\begin{figure}[h!]
 %   \centering
  %  \includegraphics[width=7cm]{images/alpha_mobilenet}
   % \caption{TPR/FPR curve for the different MobileNetV2 alternatives}
    %\label{fig:tpr_fpr_curve}
%\end{figure}

\begin{table}[h!]
    \centering
    \begin{tabular}{lccccc}
				\hlineB{3} \\
				\textbf{Model} & \textbf{$\alpha$} & \multicolumn{1}{c}{\textbf{\begin{tabular}[c]{@{}c@{}}AP\end{tabular}}} & \textbf{\begin{tabular}[c]{@{}l@{}}Parameters\end{tabular}} & \textbf{\begin{tabular}[c]{@{}l@{}}Size \end{tabular}} \\ \hline
				A98 & 1.0 & \textbf{0.891} & 1.37M & 5.25MB \\
				A98 & 0.5 & 0.839 & 0.42M & 1.60MB\\
				A98 & 0.35 & 0.615 & 0.24M & 0.93MB\\
				B133 & 1.0 & 0.804 & 4.83M & 18.4MB \\
				C98 & 1.0 & 0.873 & 4.98M & 18.9MB\\
				\hlineB{3}\\
			%	& & & & *32-bits format
			\end{tabular}
    \caption{Performance of MobileNetV2 on FDDB relative to compression ratio $\alpha$.}
    \label{tab:alpha_mbv2}
\end{table}

Table \ref{tab:results_best} shows the comparison between the MobileNetV2 and standard methods ResNet-101 and Viola-Jones. Viola-Jones from OpenCV is used as a point of reference, with a scale factor of $1.1$, but is evaluated on greyscale images. MobileNetV2 networks still reach good accuracy with about $30\times$ fewer parameters when $\alpha=1.0$, with a drop of only $2\%$ in average precision on the FDDB color dataset, compared to the performance of the ResNet-101 trained on $640$px. Quantizing the model to Q9 for computation on a processor with Fixed-Point Integer Processing Units results in a drop of $3\%$ in precision while converting the results to $16$-bits floating-point gets almost the same performance as the $32$-bits version. The MobileNetV2 with $\alpha=0.5$ still outperforms Viola-Jones algorithm, with $100\times$ fewer parameters than ResNet-101.

\begin{table}[H]
	\centering
	%\resizebox{\linewidth}{!}{
	\begin{tabular}{lcccccc}
		\hlineB{3}\\
		\textbf{Model} & $\alpha$ & \textbf{Format} & 
		%\textbf{AP on VD} &  
		\textbf{\begin{tabular}[c]{@{}l@{}}AP\end{tabular}} & \textbf{\begin{tabular}[c]{@{}l@{}} Parameters \end{tabular}}  &\textbf{\begin{tabular}[c]{@{}l@{}}Size \end{tabular}} \\ \hline
		
		%Adaboost & - & unknown (CSEM) & - & ? & ? & - & - \\ 
		Viola-Jones & - & -   &  0.738\textsuperscript{\textdagger} & - & - \\
		ResNet-101 & - & FP32 & \textbf{0.916} & 44.5M & 170 MB \\ 
		A98 & 1.0 & FP32 & 0.891 & 1.37M & 5.3 MB \\ 
		A98 & 1.0 & FP16 & 0.891 & 1.37M & 2.7 MB \\ 
		A98 & 1.0 & Q9 & 0.886 &  1.37M & 2.7 MB \\  
		A98 & 0.5 & FP32 & 0.839 &  \textbf{0.42M} & \textbf{1.60 MB} \\
		\hlineB{3}
		%\multicolumn{1}{l}{}
		%\multicolumn{1}{l}{* from OpenCV} & 
		%\multicolumn{2}{l}{* 640px, RGB+Rotation} & % 
		\multicolumn{3}{l}{\textsuperscript{\textdagger} evaluated on greyscale images}
		\vspace{0.05cm}
	\end{tabular}	
%}
\caption[Comparison of the models]{Comparison of the models in term of performance and size.}
\label{tab:results_best}
\end{table}

%% file: tex/k210.tex
\subsection{Embedding on MAIX Dock}
\label{sec:embedding}

In order to show that our model can be efficiently run on an embedded device, we use the MAIX Dock board, which is a development kit for the Kendryte K210 processor. Our implementation relies on MaixPy 4.0.13 firmware. The MobileNetV2 model A98 with $\alpha=0.5$ was chosen because of the memory limitations of the processor. The network's input is fixed to $224\times224\times3$, as bigger input images produce intermediate feature maps that do not fit in memory. The model is then converted to Tensorflow-Lite format using TensorFlow 1.13, which contains $8$-bits operations with automatic quantization and dequantization. The input and output are still in float format. We further converted the Tensorflow-Lite model to the proprietary KModel format using the MAIX Toolbox and ran it in MicroPython.

Some precautions are necessary before converting the model: the \textit{padding} operation in TensorFlow pads the input at the bottom and the right. In the MobileNetV2 model, padding occurs before depthwise-convolutions with \textit{stride} of $2$ and 'same' \textit{padding} have to be changed to a full padding (top,left,bottom,right). The depthwise-convolution padding must be set to 'valid'. These parameters can be easily changed in the \textit{json} file defining a Keras model. Once the architecture contains the right padding, it can be transformed into a KModel and uploaded to the Flash memory of the board.%, following the pipeline described in \ref{fig:pipeline_kmodel}.

%\begin{figure}[h]
%	\centering
%	\includegraphics[height=10cm]{images/kmodel_pipeline}
%	\captionsetup{justification=centering}
%	\caption[Conversion of the Keras model to KModel format]{Conversion of the Keras model to Kmodel format}
%	\label{fig:pipeline_kmodel}
%\end{figure}

\begin{figure}[ht]
	\centering
	\subfigure{\includegraphics[height=2.3cm, keepaspectratio]{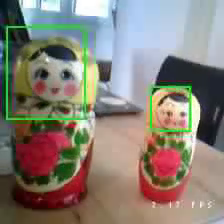}}
	\hspace{0.1cm}
	%\subfloat {\includegraphics[height=3.0cm, keepaspectratio]{images/prince}}
	%\hspace{0.5cm}
	\subfigure{\includegraphics[height=2.3cm, keepaspectratio]{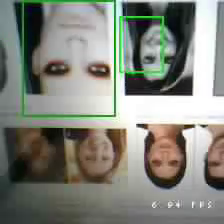}}
	\hspace{0.1cm}
	%\subfloat {\includegraphics[height=3.0cm, keepaspectratio]{images/myface2}}
	%\\ 
	%\subfloat {\includegraphics[height=3.0cm, keepaspectratio]{images/myface5}}
	%\hspace{0.5cm}
	%\subfloat {\includegraphics[height=3.0cm, keepaspectratio]{images/myface3}}
	%\hspace{0.5cm}
	\subfigure{\includegraphics[height=2.3cm, keepaspectratio]{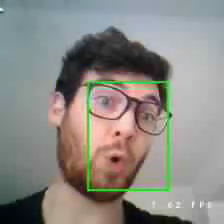}}
	%\hspace{0.5cm}
	%\subfloat {\includegraphics[height=3.0cm, keepaspectratio]{images/myface1}}
	\\
	\subfigure{\includegraphics[height=2.3cm, keepaspectratio]{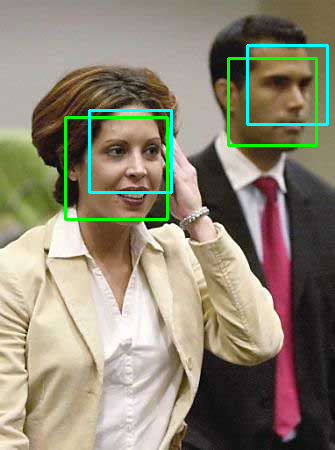}}
	\hspace{0.1cm}
	\subfigure{\includegraphics[height=2.3cm, keepaspectratio]{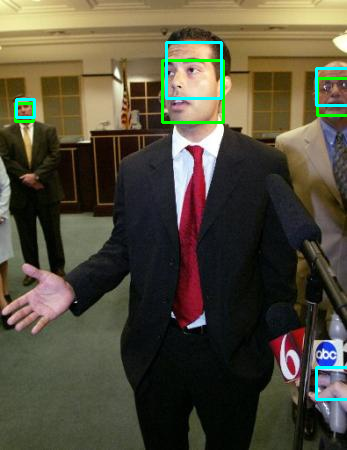}}
	\hspace{0.1cm}
	\subfigure{\includegraphics[height=2.3cm, keepaspectratio]{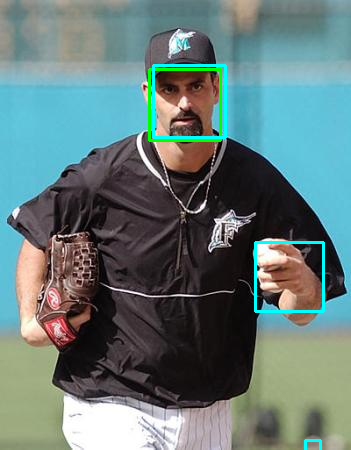}}
	
	\caption{Face detection on images from a MobileNetV2-A98. Top: online predictions of the model running on MAIX Dock.
	Bottom: Sample predicted images from the FDDB dataset by A98 with $\alpha=0.5$, offline. Cyan bounding boxes are the groundtruth, while green boxes are predicted.}
	\label{fig:kendryte_results}
\end{figure}

Qualitative results from the MAIX Dock are displayed in Figure \ref{fig:kendryte_results}. The model runs at $8$ FPS and can detect multiple faces. Because of the limitation of the image size and the available storage, FDDB could not be tested on the device, but we display offline predictions of the model at the bottom of Figure \ref{fig:kendryte_results}.

%% file: tex/conclusion.tex
\section{Conclusion}
\label{sec:conclusion}
Based on the ResNet-101 backbone, the pre-trained Tiny-Face architecture performs very well on natural images, but its size prevents it from being embedded on edge devices. Our experiments show that MobileNetV2 can successfully be applied as an alternative to classical machine learning methods for embedded face detection, as its architecture allows for a continuous trade-off between accuracy and size. This is particularly useful in certain embedded applications, where certain trade-offs such as face detection in a tighter angular or scale range can be made.

% As MobileNetV2 is constructed as a sequence of convolutional blocks, the network can be embedded quite easily, provided that the target platform contains a sufficient memory capacity for each block. 

Quantization in 16-bits format is a first step to reduce the size of the network parameters by half without losing too much precision. Quantization in Q9-format, on the other hand, allows the network to be embedded on devices that do not support floating-point operations, and achieves a similar performance as the floating-point baseline while requiring only half of its size for storing the parameters.

Future work includes assessing the network's performance on the WIDER FACE test and validations partitions, offering a large degree of variations in terms of poses and illuminations. This can provide deeper insights into the model's strengths and weaknesses. Other directions of interest include the study of early-exit strategies on other network architectures and datasets, and of modern adaptive quantization schemes. Of particular interest are  approaches that could allow dynamic adjustment of the width multiplier of the network depending on the computational resources of a platform.

% as they would provide an interesting trade-off between the desired performance and the memory footprint of the model.

%% file: biblio.tex
%\newpage
\bibliographystyle{ieeetr}
\bibliography{bibliography}
\addcontentsline{toc}{chapter}{Bibliography}